# Investigation of Frame Differences as Motion Cues for Video Object Segmentation

Sota Kawamura[1][0009-0006-5597-076X]*, Hirotada Honda[1], Shugo Nakamura[1], Takashi Sano[1]**

[1] Toyo University, 115-0053 Tokyo, Japan
```
*s3F102300056@iniad.org,**takashi.sano@iniad.org
```

**Abstract.** Automatic Video Object Segmentation (AVOS) refers to the task of autonomously segmenting target objects in video sequences without relying on human-provided annotations in the first frames. In AVOS, the use of motion information is crucial, with optical flow being a commonly employed method for capturing motion cues. However, the computation of optical flow is resource-intensive, making it unsuitable for real-time applications, especially on edge devices with limited computational resources. In this study, we propose using frame differences as an alternative to optical flow for motion cue extraction. We developed an extended U-Net-like AVOS model that takes a frame on which segmentation is performed and a frame difference as inputs, and outputs an estimated segmentation map. Our experimental results demonstrate that the proposed model achieves performance comparable to the model with optical flow as an input, particularly when applied to videos captured by stationary cameras. Our results suggest the usefulness of employing frame differences as motion cues in cases with limited computational resources.

## 1 INTRODUCTION

Segmentation is the task of identifying the location of target objects within an image. It can be applied to both images and videos, with applications, for example, in environmental recognition for autonomous vehicles, dynamic scene analysis for surveillance systems, and automated video editing. Advances in machine learning, particularly deep learning, have played a critical role in the development of these techniques.

Automatic Video Object Segmentation (AVOS) is a process that identifies and segments objects in videos without the need for prior information. Many previous studies [1, 7, 14] have employed optical flow to capture the motion information of objects. Optical flow, which estimates the direction and velocity of pixel movement, plays a crucial role in enhancing segmentation accuracy. However, despite its effectiveness, the computation of optical flow is resourceintensive, limiting its applicability in real-time scenarios, particularly on resource-constrained devices.

To address the limitations of optical flow, we propose an alternative approach that utilizes pixel differences between temporally adjacent frames as motion information. Frame differences is computationally efficient and captures changes in pixel intensity between consecutive frames, providing a simpler method for detecting motion cues. Although frame differences has been previously explored in the context of segmentation [6], its integration into modern deep learning-based methods has not been widely adopted. In this study, we develop a segmentation model based on an extended ResNet [2], which incorporates frame differences into the video object segmentation process. We evaluate the impact of this approach on segmentation performance, particularly in comparison to optical flow-based methods.

The remainder of this paper is structured as follows. Section 2 reviews related work, emphasizing prior approaches to video segmentation and motion estimation. In Section 3, we describe the proposed method. Section 4 presents the experimental setup and results, including a comparative analysis of segmentation performance. Finally, Section 5 is devoted to the summary and discussion.

## 2      RELATED WORK

Several deep learning-based models have been applied to AVOS tasks [1, 4, 7, 14]. These models use optical flow or similar motion estimation methods as motion information. Considering the computational cost of estimating optical flow, running these models in real-time on edge devices, such as smartphones, is challenging. While general-purpose video segmentation models, like the Segmentation Anything Model (SAM) [11], also exist, their transformer architecture requires significantly more memory compared to CNNs, making them similarly unsuitable for resource-limited environments.

Optical flow represents a vector field indicating the direction and magnitude of pixel movement in video frames [5]. Typically, optical flow is computed by solving an optimization problem [3]. However, since this approach requires iterative optimization of complex cost functions, it is difficult to achieve real-time processing. More recently, deep learning methods have been proposed for estimating optical flow, such as PWC-Net [13]. These methods enable fast estimation of optical flow when sufficient computational resources, like GPUs, are available. Nevertheless, these techniques are not well-suited for use cases on edge devices with limited computational power.

In videos, pixel intensity differences between frames can be considered as conveying motion information. The application of frame differences for video segmentation was explored in the earlier study [6]. However, there has been little research on whether frame difference can serve as a useful feature for segmentation in modern deep learning-based methods, especially in comparison to optical flow. With the rise of deep learning, methods like ResNet [2] have not yet fully explored the potential of frame differences as an input for video object segmentation.

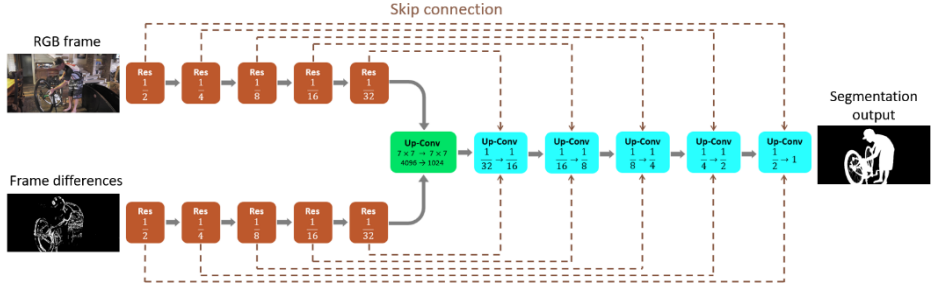

**Fig. 1.** Overview of the proposed dual-encoder architecture for Automatic Video Object Segmentation (AVOS). The model employs two encoders based on ResNet50, one for RGB video frames and the other for frame difference inputs. Skip connections convey features from each encoder to the corresponding layers in the decoder. The decoder, consisting of upsampling and convolutional layers, outputs a segmentation map matching the resolution of the input frame. The use of frame differences allows the model to capture motion information efficiently.

## 3    METHOD

We first define a frame difference from a video or a sequence of images. Let $X_t$ be an image in a frame at time $t$. We define the frame difference as:

$$X_t^{\text{diff}} = \text{abs}(X_{t+1} - X_t), \qquad (1)$$

where $\text{abs}(\cdot)$ represents element-wise absolute values. The absolute value is taken because we focus on the motion between pixels, regardless of object appearance or disappearance.

To investigate the effectiveness of frame differences in AVOS tasks, we employ a fully convolutional neural network similar to U-Net[12]. As shown in Fig. 1, the proposed model consists of two encoders and one decoder.

The two encoders have the same architecture as ResNet50[2], where we use publicly available pre-trained parameters on the ImageNet dataset. A video frame $X_t$, along with the frame difference $X_t^{\text{diff}}$, are fed into the two encoders, respectively, to obtain feature maps. The encoder outputs are multi-scale feature maps from five residual blocks, which are forwarded to the decoder through skip connections to preserve object details. The bottleneck of the network combines the outputs of the two encoders. The feature maps from the deepest layers of both encoders are concatenated and passed through a `1 x 1` convolution layer to reduce their dimensionality to `1024` channels. This step integrates both motion information (captured by $X_t^{\text{diff}}$) and appearance information (captured by $X_t$).

The decoder reconstructs the segmentation map from the bottleneck features by sequentially performing upsampling and convolution operations. At each upsampling stage, the decoder concatenates features from the corresponding residual blocks of the encoders via skip connections. This hierarchical design ensures that both global context and local spatial details are preserved in the segmentation map. The overall mod-el is trained in a supervised manner by minimizing the cross-entropy loss with respect to the annotated segmentation maps.

In order to evaluate the effectiveness of frame differences, we additionally consider two similar models for comparison. One model takes the video frame and the corresponding optical flow instead of the frame difference, while the other one takes only the video frame with no motion cues fed into the model.

## 4      EXPERIMENTS AND RESULTS

To evaluate the effectiveness of the proposed method, we utilize the DAVIS[10] and FBMS-59[9] datasets, which are widely recognized in video segmentation research. The DAVIS dataset provides high-quality annotations across various scenarios, making it a standard benchmark for evaluating segmentation models. Meanwhile, the FBMS-59 dataset contains a variety of complex scenes and object movements, offering a suitable testbed for assessing the model's generalizability across different conditions.

The model performance is evaluated using two primary metrics: contour accuracy $F$ and region similarity (Intersection over Union, IoU). Contour accuracy $F$ is the F-measure for the binary classification of the segmentation boundary's correctness at each pixel Region similarity is defined as:

$$IoU = \frac{|M \cap G|}{|M \cup G|},\qquad(2)$$

where $M$ represents the predicted segmentation map and $G$ is the ground truth. This metric quantifies the overlap between the estimated segmentation and the ground-truth segmentation.

Firstly, we used all the training data from the DAVIS dataset for training, aiming to cover a wide range of scenes. The three models were trained five times with different initial conditions to account for variability. We employed cross entropy as the loss function and used the Adam[8] algorithm for optimization. The learning rate was set to 0.005, and training was conducted for 100 epochs with a mini-batch size of 16.

**Table 1.** Average contour accuracy (F) and intersection-over-union (IoU) of three models (Proposed Method, Optical Flow, No Motion Information) on the DAVIS and FBMS-59 datasets. Each model was trained five times under the same conditions.

| Condition | Metric | Proposed Method | Optical Flow | No Motion |
|---|---|---|---|---|
| DAVIS | F | 0.690 | 0.714 | 0.595 |
|  | IoU | 0.623 | 0.642 | 0.518 |
| FBMS-59 | F | 0.549 | 0.634 | 0.548 |
|  | IoU | 0.124 | 0.110 | 0.123 |

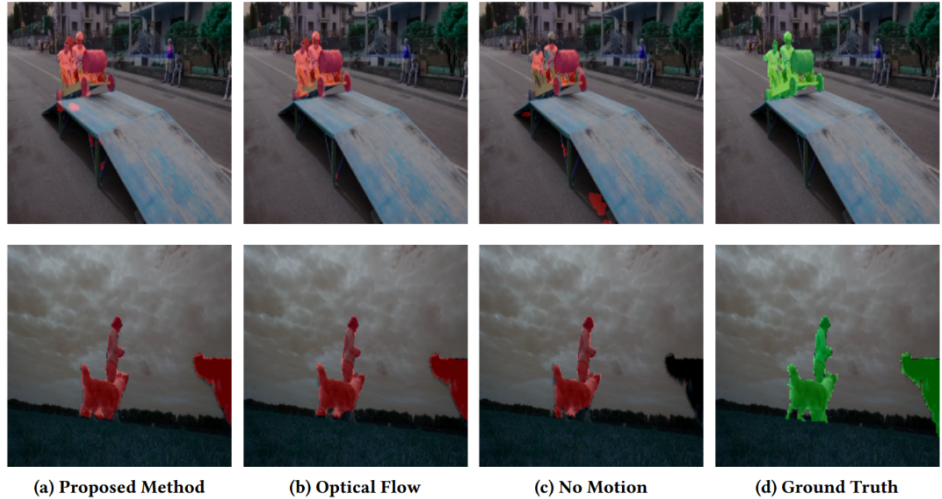

(a) Proposed Method  (b) Optical Flow  (c) No Motion  (d) Ground Truth

**Fig. 2.** Examples of predicted segmentation results from three models (Proposed Method, Optical Flow, No Motion Information) trained on the entire DAVIS dataset.

After training, we evaluated the contour accuracy ($F$) and region similarity (IoU) on the test datasets of DAVIS and FBMS-59. The results are summarized in Table 1. As shown in Table 1, while the model using optical flow achieved the best performance, the model utilizing frame differences also demonstrated comparable results. For FBMS-59, IoU values are small for all methods. This is likely due to the difference in data characteristics between DAVIS and FBMS59, as the training was carried out on the DAVIS dataset. Fig. 2 presents examples of the predicted segmentation results. Secondly, we examined the performance under conditions where frame differences was expected to be particularly effective for segmentation. Specifically, we hypothesized that the effectiveness of frame differences would be maximized in scenarios where the camera remains stationary and the target objects are in motion. Therefore, we selected 12 videos from the DAVIS dataset, which had been captured by stationary cameras, for model training and evaluation. The evaluation was conducted using four-fold cross-validation, and the FBMS59 dataset was also used to test generalizability. The results are summarized in Table 2. In this case, we found that the model using frame differences achieved the highest performance in the DAVIS dataset. We observe that IoU values for FBMS-59 are small for all methods again, owing to the different characteristics between DAVIS and FBMS-59. Fig. 3 shows examples of the predicted results from the DAVIS dataset. Both quantitative and qualitative evaluations confirmed that segmentation accuracy was improved when frame differences were used, compared to models combining original images with optical flow.

**Table 2.** Contour accuracy (F) and intersection-over-union (IoU) of three models (Proposed Method, Optical Flow, No Motion Information) evaluated on stationary camera footage from the DAVIS and FBMS-59 datasets, using four-fold cross-validation.

| Condition | Metric | Proposed Method | Optical Flow | No Motion |
|---|---|---|---|---|
| DAVIS | F | 0.658 | 0.590 | 0.525 |
| | IoU | 0.527 | 0.443 | 0.377 |
| FBMS-59 | F | 0.516 | 0.472 | 0.546 |
| | IoU | 0.040 | 0.063 | 0.068 |

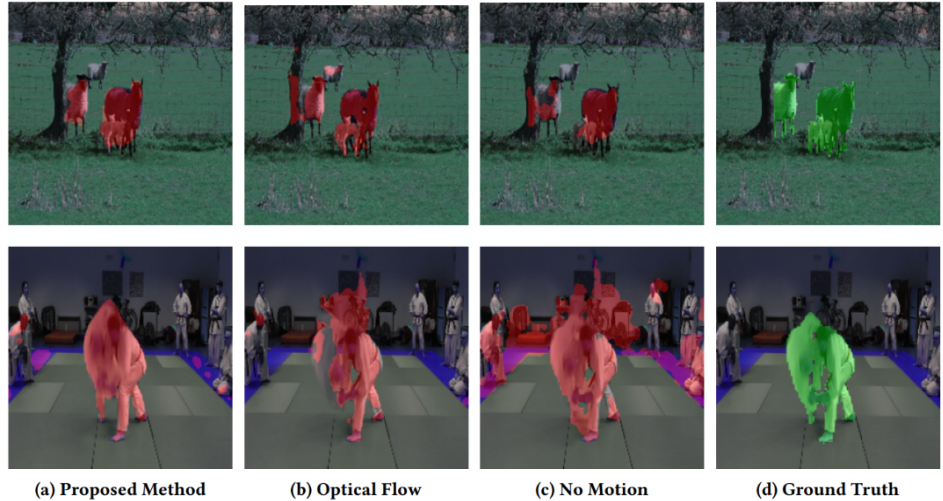

**Fig. 3.** Examples of predicted segmentation results from three models (Proposed Method, Optical Flow, No Motion Information) trained on stationary camera footage from the DAVIS dataset. The proposed method shows comparable performance to the optical flow model in terms of contour accuracy.

We discuss the computational efficiency of the proposed method in comparison to previous approaches. For example, according to [15], obtaining optical flow as a motion cue using PWC-Net requires approximately 0.2 seconds per frame. In contrast, frame differences, which involve only element-wise absolute differences, demand significantly fewer computational resources. This efficiency makes the proposed approach particularly advantageous for resource constrained or real-time processing scenarios.

## 5      SUMMARY AND DISCUSSION

In this study, we proposed a video segmentation method that utilizes frame differences based on pixel intensity changes. We extended a U-Net model to accept both the target segmentation frame and the frame difference as inputs, generating segmentation maps from these combined features.

When the training data consisted of videos captured by stationary cameras, the proposed method outperformed not only the model without motion information but also the model using optical flow. This suggests that frame differences effectively capture motion information in such scenarios.

However, when the entire DAVIS dataset, which includes videos with non-stationary cameras, was used, the model incorporating optical flow achieved the highest accuracy. While optical flow offers robust performance across a wide variety of scenes, the frame differences model produced nearly comparable results, demonstrating that frame differences can approach the performance of optical flow-based models despite being computationally more efficient. This highlights the potential of frame differences as a viable alternative to optical flow, particularly in resource-constrained environments.

Despite its computational simplicity, frame differences were found to improve segmentation results similarly to optical flow. This result is intuitively reasonable, but the relationship between optical flow and frame differences can be understood more formally as follows: Let $I(x, y, t)$ represent the brightness of a pixel $(x, y)$ at time frame $t$. The optical flow $(V_x, V_y)$ at pixel $(x, y)$ and time $t$ satisfies the following equation:

$$I_x(x,y,t)V_x + I_y(x,y,t)V_y = -I_t(x,y,t), \tag{3}$$

where $I_x(x,y,t)$, $I_y(x,y,t)$, and $I_t(x,y,t)$ are the partial derivatives of $I(x,y,t)$ with respect to $x$, $y$, and $t$, respectively. In other words, the dot product of the optical flow and the spatial derivatives is related to the temporal derivative. In the case of discrete time, the temporal derivative on the right-hand side becomes the frame difference. Therefore, using frame differences (the right-hand side) is equivalent to using the dot product of the brightness spatial derivatives with the optical flow (the left-hand side). While it is generally impossible to reconstruct the optical flow, which is a velocity vector field, from frame differences alone, some information carried by the optical flow is expected to also be present within the frame differences.

Although this study employed a U-Net-like architecture, the proposed approach is not limited to this model. Frame differences could be integrated into more advanced models as an alternative to optical flow. Future work will explore the potential of frame differences further, as well as the trade-offs between computational cost and segmentation accuracy.

**Acknowledgments.** This work was supported by Toyo University Top Priority Research Program.

**Disclosure of Interests.** The authors have no competing interests to declare that are relevant to the content of this article.